\documentclass[runningheads]{llncs}
\usepackage{graphicx}
\usepackage{amsmath,amssymb} 
\usepackage{color}
\usepackage{array}
\usepackage{multirow}
\usepackage{booktabs}
\usepackage{amssymb} 
\usepackage{hyperref}
\begin{document}
\title{RedNet: Residual Encoder-Decoder Network for indoor RGB-D Semantic Segmentation}
\titlerunning{RedNet for indoor RGB-D Semantic Segmentation}
%
\author{Jindong Jiang \and
Lunan Zheng \and
Fei Luo \and Zhijun Zhang}
\authorrunning{J. Jiang et al.}
%

\institute{The School of Automation Science and Engineering, South China University of Technology, Guangzhou 510640, China \\ \email{jdpshq@gmail.com, aulnzheng@sina.com, \\ \{aufeiluo,auzjzhang\}@scut.edu.cn}}

\maketitle              
\begin{abstract}
Indoor semantic segmentation has always been a difficult task in computer vision. In this paper, we propose an RGB-D residual encoder-decoder architecture, named RedNet, for indoor RGB-D semantic segmentation. In RedNet, the residual module is applied to both the encoder and decoder as the basic building block, and the skip-connection is used to bypass the spatial feature between the encoder and decoder. In order to incorporate the depth information of the scene, a fusion structure is constructed, which makes inference on RGB image and depth image separately, and fuses their features over several layers. In order to efficiently optimize the network's parameters, we propose a `pyramid supervision' training scheme, which applies supervised learning over different layers in the decoder, to cope with the problem of gradients vanishing. Experiment results show that the proposed RedNet(ResNet-50) achieves a state-of-the-art mIoU accuracy of 47.8\% on the SUN RGB-D benchmark dataset.

\end{abstract}
\section{Introduction}

Indoor space is likely to be the main workplace for service robots in the near future. In order to work well in an indoor space, the robots should possesses the ability of visual scene understanding. To do so, the semantic segmentation in indoor scene is becoming one of the most popular tasks in computer vision.

Over the pass few years, fully convolutional networks (FCNs) type architectures have shown great potential on semantic segmentation task \cite{long2015fully,noh2015learning,badrinarayanan2015segnet,chen2016deeplab,yu2015multi,lin2017refinenet,yu2017dilated}, and have dominated the semantic segmentation task of many datasets \cite{everingham2010pascal,cordts2016cityscapes,song2015sun}. Some of this FCNs-type architectures focus on indoor environment, and usually utilize the depth information as the complementary information for RGB to improve the segmentation \cite{long2015fully,couprie2013indoor,gupta2014learning,hazirbas2016fusenet}. In general, the FCNs architectures can be generally divide into two categories, i.e., the encoder-decoder type architectures and dilated convolution architectures. The encoder-decoder architectures \cite{long2015fully,noh2015learning,badrinarayanan2015segnet,lin2017refinenet,hazirbas2016fusenet} have a downsample path to extract the semantic information from images and a upsample path to recover a full-resolution semantic segmentation map. By contrast, the dilated convolution architectures \cite{chen2016deeplab,yu2015multi,yu2017dilated} employ dilated convolution such that the convolutional network expands receptive field exponentially without downsampling. With less or even zero downsampling operation, dilated architectures keep the spatial information in the image through out the whole networks, so the architectures serve as a discriminative model that classify every pixel on the image. Encoder-decoder architectures, on the other hand, lost spatial information during the discriminative encoder, and thus some of the networks apply skip-architecture to recover the spatial information during the generative decoder path.

Even though the dilated convolution architectures have the advantage of keeping the spatial information, they generally have higher memory consumption on the training step. Because the spatial resolution of the activation map is not downsampled as the network proceed and it needs to be stored for gradient computation. Therefore, the high memory consumption stops the network from having a deeper structure. This could cause disadvantages on this method, since convolutional networks learn richer features as the structure gets deeper, which would benefit the inference of the semantic information. 

\begin{figure}[!t]
    \centering
    \includegraphics[width=0.85\linewidth]{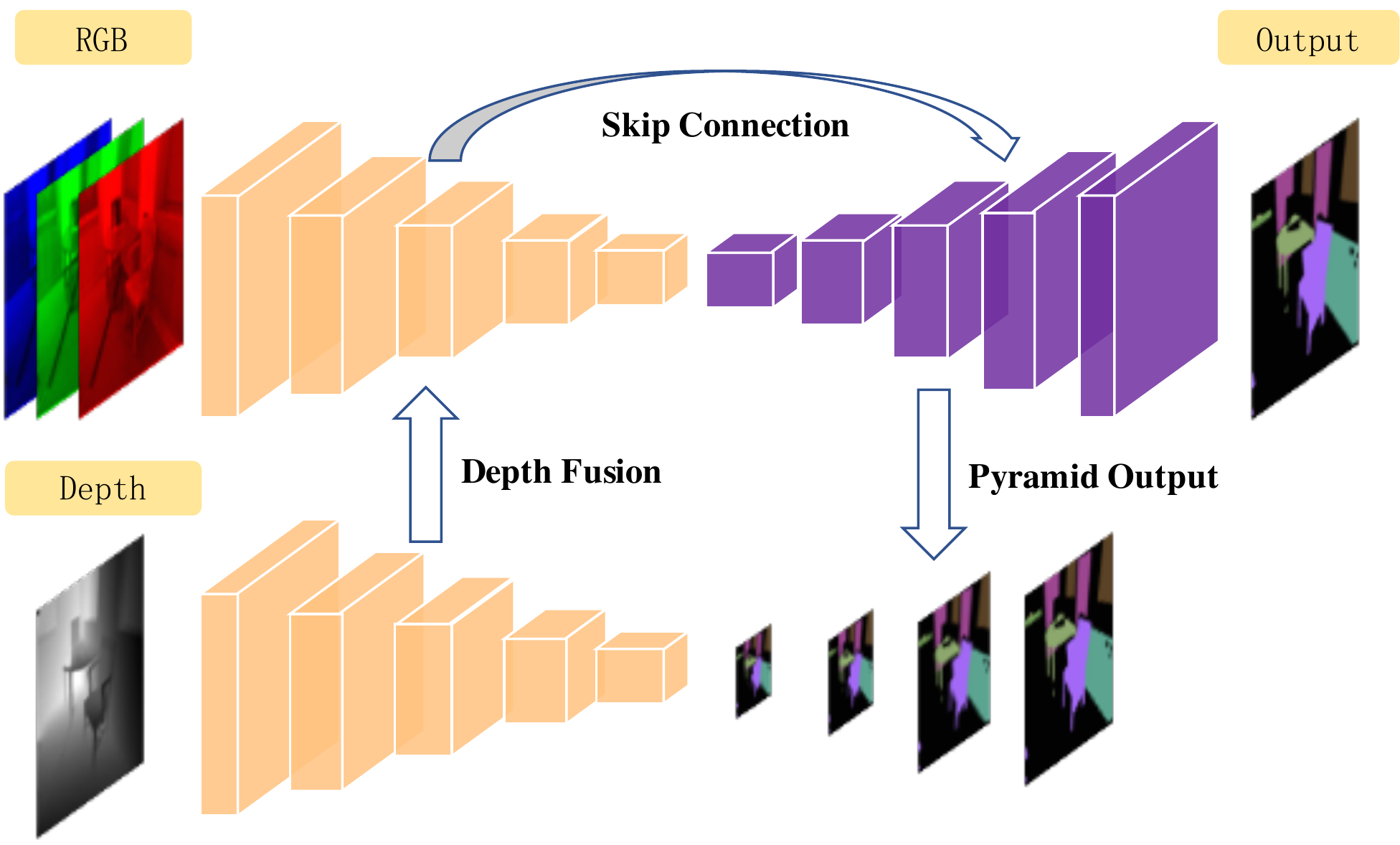}
    \caption{Overall structure of the proposed network.}
    \label{fig:Overall}
\end{figure}

In this paper, we propose a novel structure named RedNet that employ the encoder-decoder network structure for indoor RGB-D semantic segmentation. In RedNet, the residual block is used as the building module to avoid the model degradation problem \cite{he2016deep}. This allows the performance of networks to improve as the structure goes deeper. Moreover, we apply fusion structure to incorporate depth information into the network, and use skip-architecture to bypass the spatial information from encoder to decoder. Further, inspired by the training scheme in \cite{szegedy2015going}, we propose the pyramid supervision that apply supervised learning over different layers on the decoder for better optimization. The overall structure of RedNet is illustrated in Fig. \ref{fig:Overall}.

The remainder of this paper is organized in four sections. In section \ref{sec:relatedwork}, the literature on residual networks and indoor RGB-D semantic segmentation is previewed. The architecture of RedNet and the idea of pyramid supervision are stated in detail in section \ref{sec:approach}. In section \ref{sec:experiment}, the comparative experiments are conducted to evaluate the efficiency of the model. Finally, we draw a conclusion of this paper in section \ref{sec:conclusion}.

Before ending this section, the main contributions of this paper are listed as the following.
\begin{enumerate}
    \item[1.] A novel residual encoder-decoder architecture (termed RedNet) is proposed for indoor RGB-D semantic segmentation, which applies residual module as the basic building block in both the encoder path and decoder path.
    \item[2.] A pyramid supervision training scheme is proposed to optimize the network, which applies supervised learning over different layers on the upsample path of the model.
    \item[3.] Two comparison experiments are conducted on SUN RGB-D benchmark to verify the effectiveness of the proposed RedNet architecture and the pyramid supervision training scheme.
\end{enumerate}

\section{Related Work} \label{sec:relatedwork}

\subsection{Residual Networks}

Residual network was first proposed by He et al. in \cite{he2016deep}. In their work, they analyzed the problem of model degradation, which present as saturation and then degradation of accuracy as the network depth increasing. They argued that the degradation problem is an optimization problem, and as the depth of the network increase, the network gets harder to train. It was assumed that the desired mapping of a convnet is comprised of an identity mapping and a residual mapping. Therefore, a deep residual learning framework is proposed. Instead of letting a convnet learn the desired mapping, it fits the residual mapping and uses shortcut connection to merge it with the identity input. With this configuration, the residual network become easy to optimize and can enjoy accuracy gains from greatly increased depth. Veit et al. \cite{veit2016residual} presented a complementary explanation of the increased performance of residual networks, i.e., the residual networks avoid the vanishing gradient problem by introducing the short paths between input and output. Later, He et al. \cite{he2016identity} analyzed the propagation formulations behind the connection mechanisms of residual networks and proposed a new structure of residual unit. In their work, they extended the depth of a deep residual networks to 1001 layers.  Zagoruyko et al. \cite{zagoruyko2016wide} investigated the memory consumption of residual networks and propose a novel residual unit that aims to decrease depth and increase width of a deep residual network.

The idea of residual learning was later adopted to architectures for semantic segmentation task. Pohlen et al. \cite{pohlen2017full} proposed a fully convolutional network with residual learning for semantic segmentation in street scenes. The network has an encoder-decoder architecture and applies residual module on the skip-connection structure with the full-resolution residual units (FRRUs). Quan et al. \cite{quan2016fusionnet} presented a FCN architecture, named FusionNet, for connectomics image segmentation. Instead of using residual block on skip-connection structure, FusionNet applies them on each layer in the encoder and decoder path along with standard convolution, max-pooling, and transpose of convolution \cite{noh2015learning}. Similarly, Drozdzal et al. \cite{drozdzal2016importance} studied the importance of skip-connection in biomedical image segmentation, showing that the ``short skip connections'' in residual module is more effective than the ``long skip connections'' between encoder and decoder on biomedical image analyzing. Yu et al. \cite{yu2017dilated} combined the idea of residual networks and dilated convolution to build a dilated residual networks for semantic segmentation. In their paper, they also studied the gridding artifact introduced by dilation convolution and developed a `degridding' method to removing these artifacts. Dai et al. \cite{dai2016instance} used ResNet-101 as the basic network and apply the Multi-task Network Cascades for instance segmentation. Lin et al. \cite{lin2017refinenet} and Lin et al. \cite{lin2017cascaded} also used ResNet structure as a feature extractor and employed a multi-path refinement network to exploits information along the down-sampling process for full resolution semantic segmentation.

In 2017, Chaurasia et al. \cite{chaurasia2017linknet} proposed a encoder-decoder architecture (named LinkNet) for efficient semantic segmentation. The LinkNet architecture uses ResNet18 as the encoder and applies the bottleneck unit in the decoder for feature upsample. Under this efficient configuration, the network achieve state-of-the-art accuracy on several uban street dataset \cite{cordts2016cityscapes,brostow2008segmentation}. Inspired by this work, we propose a straightforward encoder-decoder structure that apply residual unit on both the downsample path and upsample path, and employs the pyramid supervision to optimize it.

\subsection{Indoor RGB-D Semantic Segmentation}

Currently, the accurate indoor semantic segmentation is still a challenging problem due to the high similarity of color and structure between objects, and the non-uniform illumination in indoor environment. Therefore, some work started utilizing the depth information as the complementary information to solve the problem. For instance, Koppula et al. \cite{koppula2011semantic} and Huang et al. \cite{huang2014object} used depth information to build 3D point clouds of full indoor scenes, and applied graphical model to capture features and contextual relations of objects in RGB-D data for semantic labeling. Gupta et al. \cite{gupta2013perceptual} proposed a superpixel-based architecture for RGB-D semantic segmentation in indoor scene. Their method applied superpixel regions extraction on RGB image and feature extraction of each superpixel on RGB-D data, then employ Random Forest (RF) and Support Vector Machine (SVM) to classify each superpixel and build a full-resolution semantic map. Later, Gupta et al. \cite{gupta2014learning} improved this segmentation model by introducing a HHA encoding for depth information and use a Convolutional Neural Network (CNN) for feature extraction. In HHA encoding, depth information is encoded into three channel, i.e., horizontal disparity, height above ground, and angle between gravity \& surface normal. These implied that the HHA encoding emphasize the geocentric discontinuities in the image.

After the release of several indoor RGB-D datasets \cite{silberman2011indoor,janoch2011category,silberman2012indoor,song2015sun}, many researches started employing deep learning architectures for indoor semantic segmentation. Couprie et al. \cite{couprie2013indoor} presented a multi-scale convolutional network for indoor semantic segmentation. The study showed that the recognition of object classes with similar depth appearance and location is improved when incorporating the depth information. Long et al. \cite{long2015fully} applied FCNs structure on indoor semantic segmentation and compare different inputs to the network, including three channel RGB, stacked four channel RGB-D, and stacked six channel RGB-HHA. The research further showed that the RGB-HHA input outperform all other input form, while the RGB-D have similar accuracy with RGB input. Hazirbas et al. \cite{hazirbas2016fusenet} presented a fusion-based encoder-decoder FCNs for indoor RGB-D semantic segmentation. Their work shows that the HHA encoding does not hold more information than the depth itself. In order to fully utilize the depth information, they apply two branches of convolutional network to compute RGB and depth image respectively and apply features fusion on different layers. Based on the same depth fusion structure, our previous work \cite{jiang2017incorporating} proposed a DeepLab-type architecture \cite{chen2016deeplab} that applies depth incorporation on a dilated FCNs and build a RGB-D conditional random field (CRF) as the post-process.

In this work, we will also apply depth fusion structure on the downsample part of the network, and apply skip-connection to bypass the fused information to the decoder for full-resolution semantic prediction.

\section{Approach} \label{sec:approach}

\subsection{RedNet Architecture}

\begin{figure}[!t]
    \centering
    \includegraphics[width=0.70\linewidth]{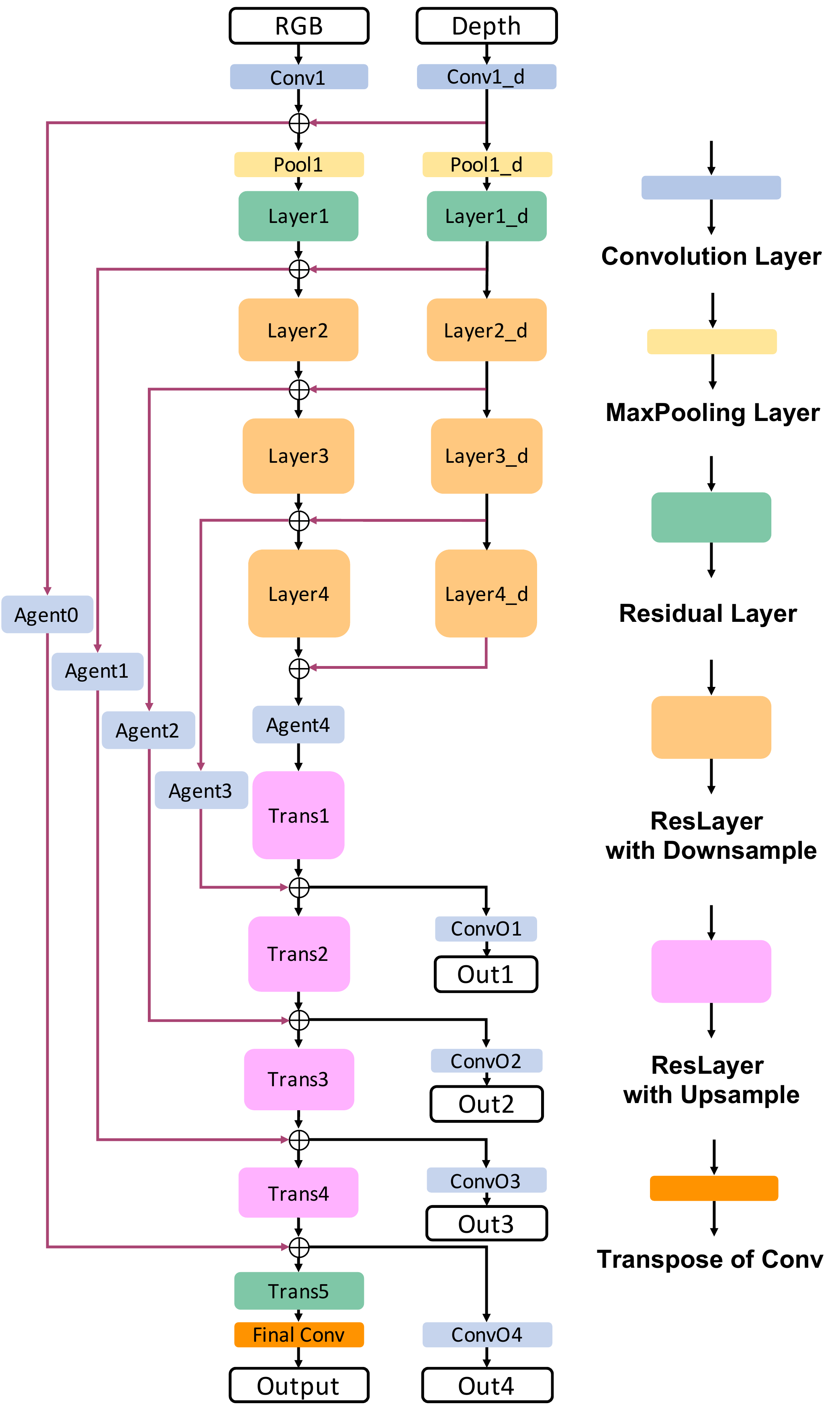}
    \caption{Layer configuration of the proposed RedNet (ResNet-50).}
    \label{fig:rednet}
\end{figure}

The architecture of RedNet is presented in Fig. \ref{fig:rednet}. For clear illustration, we use blocks with different color to indicate different kinds of layers. Notice that each convolution operation in RedNet is followed by a batch normalization layer \cite{ioffe2015batch} before relu function, and it is omitted in the figure for simplification.

The upper half of the figure up to Layer4/Layer4\_d is the encoder of the network, it has two convolutional branches, i.e., the RGB branch and the Depth branch. Structures of both encoder branches can be adopted from one of the five ResNet architectures proposed in \cite{he2016deep}, in which we remove the last two layers of ResNet, i.e., the global average pooling layer and fully-connected layer. The RGB branch and the Depth branch in the model have the same network configuration, except that the convolution kernel of Conv1\_d on Depth branch has only one feature channel, since the Depth input presented as an one channel gray image. The encoder starts with two downsample operation, which is the \(7 \times 7\) convolution layer with stride two and a \(3 \times 3\) max-pooling layer with stride two. This max-pooling is the only pooling layer in the whole architecture, all other downsample and upsample operations in the network are implemented with two-stride convolution and transpose of convolution. The following layers in encoder are residual layers with different numbers of residual unit. It is worth pointing out that only Layer1 in the encoder does not have downsample unit, and all other ResLayer have one residual unit that downsample the feature map and increase the feature channel by a factor of 2. The Depth branch ending at Layer4\_d, and its features are fused into RGB branch on five layers. Here, element-wise summation is performed as the feature fusion method.

\begin{figure}[!t]
    \centering
    \includegraphics[width=0.95\linewidth]{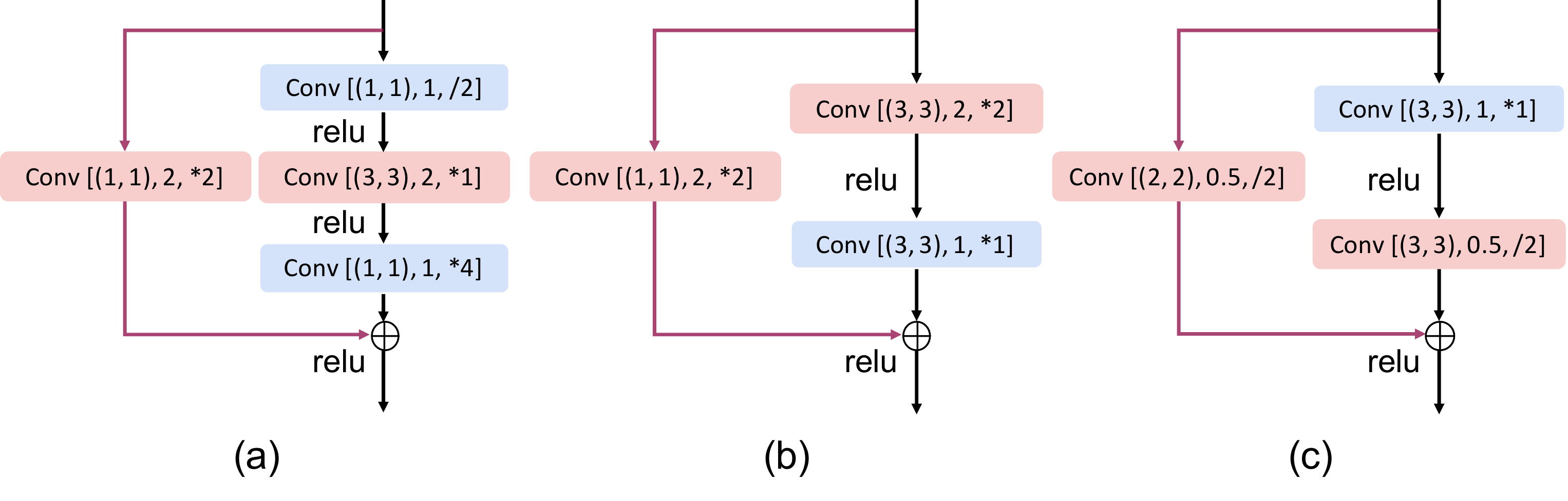}
    \caption{Downsample and upsample residual unit. (a): a downsample residual unit in (ResNet-50) encoder. (b): a downsample residual unit in (ResNet-34) encoder. (c): a upsample residual unit we propose in decoder.}
    \label{fig:encoder_decoder_unit}
\end{figure}

The lower half of Fig. \ref{fig:rednet}, starting with Trans1 layer, is the decoder of the network. Here, except the Final Conv layer, which is a single \(2 \times 2\) transpose of convolution layer, all other layers in the decoder are residual layers. The first four layers, i.e., the Trans1, Trans2, Trans3, and Trans4, have one upsample residual unit to upsample the feature map by a factor of 2. Different from the bottleneck building block in the encoder, we employ the standard residual building block \cite{he2016deep} in the decoder that have two consecutive \(3 \times 3\) convolution layers for residual computation. With regard to the upsample operation, we present a upsample residual unit that is shown in Fig. \ref{fig:encoder_decoder_unit}(c). In Fig. \ref{fig:encoder_decoder_unit}, we compare the downsample unit in ResNet-50 and ResNet-34, as well as the upsample unit we propose in the decoder. Here, for Conv\([(k,k),s,*/c]\), \((k,k)\) means the spatial size of the convolution kernel. Parameter \(s\) is the stride of the convolution, and \(c\) is the increase or decrease factor of the output feature channel. Red block denotes the convolution that changes the spatial size of the input feature map, i.e., downsample or upsample. For example, a \mbox{\(Conv[(2, 2), 0.5, /2]\)} in red means a \(2 \times 2\) kernel size transpose of convolution that upsample the width and height of the feature map by a factor of 2 and decrease the feature channel by a factor of 2.



\textbf{}

\begin{table}[!b]
    \footnotesize
    \caption{Encoder (ResNet-50) and Decoder configuration}
    \label{tab:maps_configuration}
    \centering
    \begin{tabular}{ m{3em} m{0.5em} m{2.5em} m{2.5em} m{2.5em} m{2em} m{3em} m{0.5em} m{2.5em} m{2.5em} m{2.5em} }
        \toprule
        \multirow{2}{*}{Block} & &\multicolumn{3}{c}{Encoder} & &\multirow{2}{*}{Block} & &\multicolumn{3}{c}{Decoder} \\
        \cmidrule{3-5} \cmidrule{9-11}
                & &\(m\)     &\(n\)      &\(l_\text{unit}\) & & & & \(m\) &\(n\) &\(l_\text{unit}\) \\
        \midrule
        Layer4  & &1024    &2048     &3   & &Trans1    & &512     &256   &6\\
        Layer3  & &512     &1024     &6   & &Trans2    & &256     &128   &4\\
        Layer2  & &256     &512      &4   & &Trans3    & &128     &64    &3\\
        Layer1  & &64      &256      &3   & &Trans4    & &64      &64    &3\\
        Conv1   & &3       &64       &-   & &Trans5    & &64      &64    &3\\
        \bottomrule
    \end{tabular}
\end{table}

Table \ref{tab:maps_configuration} shows the network configuration when using ResNet-50 as the encoder, here \(m\) denotes the number of input feature channel, \(n\) denote the number of output feature channel, and \(l_\text{unit}\) denote the number of residual unit in that layer. The upsample ResLayer has different residual unit order compared with the downsample ResLayer. The downsample layer starts with a downsample residual unit and followed by several residual units, by contrast, the upsample layer starts with several residual unit and ends with one upsample residual unit. As shown in the table, the output of residual layer in ResNet-50 encoder has large channel size since it use channel expansion. Therefore, we employ the \textit{Agent} layers shown in Fig. \ref{fig:rednet}, which are single \(1 \times 1\) convolutional layer with strides one. It is designed to project the feature map for lower channel size, allowing the decoder to have a lower memory consumption. Notice that the agent layers only exist when ResNet-50 is employed, they will be removed when the encoder employ ResNet-34 structure. This is because it does not have channel expansion on residual unit. In addition, we also remove skip-connection between output of Conv1 and output of Trans4 on ResNet-34 encoder setting for better performance. 

\subsection{Pyramid Supervision}
The pyramid supervision training scheme alleviate the gradient vanishing problem by introducing supervised learning over five different layers. As shown in Fig. \ref{fig:rednet}, the algorithm compute four intermediate outputs from feature maps of four upsample ResLayer in addition to the final output, these intermediate outputs are called side outputs. Each side output score map is computed using a convolution layer with \(1 \times 1\) kernel size and stride one. Therefore, all outputs have different spatial resolutions. The final \textit{Output} of RedNet is a full resolution score map, while the side outputs \textit{Out4}, \textit{Out3}, \textit{Out2}, and \textit{Out1} are downsampled. For instance, the \textit{Out1} has 1/16 the height and width of the \textit{Output}. The four side outputs and the final output are then feed into a softmax layer and cross entropy function to build the loss function. 

\begin{equation} \label{eq:each_loss}
    Loss(\mathbf{s}, \mathbf{g}) = \frac{1}{N} \sum_{i} -\log\left(\frac{\exp(s_{i}[g_{i}])}{\sum_{k} \exp(s_{i}[k])}\right)
\end{equation}

More concretely, the loss function of each output has the same form shown in Eq. \ref{eq:each_loss}. Here, \(g_i \in \mathbb{R}\) denote the class index on the groundtruth semantic map on location \(i\). \(s_i \in \mathbb{R}^{N_c}\) denote the score vector of the network output on location \(i\) with \(N_c\) being the number of classes in the dataset. \(N\) denotes the spatial resolution of the specific output. When dealing with the loss function of \(Out1\) to \(Out4\), the groundtruth map \(\mathbf{g}\) is downsampled using nearest-neighbor interpolation. The overall cross entropy loss is thus the summation of all five cross entropy losses over five outputs. Notice that instead of assigning equally-weighted loss on pixels in different outputs, these overall loss configuration assign more weight on pixels of downsampled output, e.g., \textit{Out1}. In practice, we find that this configuration provide better performance than the equally-weighted loss configuration. 

\section{Experiment} \label{sec:experiment}

In this section, we evaluate the RedNet architectures with ResNet-34 and ResNet-50 as the encoder using the SUN RGB-D indoor scene understanding benchmark suit \cite{song2015sun}. The SUN RGB-D dataset is currently the largest RGB-D indoor scene semantic segmentation dataset It has 10,335 densely annotated RGB-D images taken from 20 different scenes, at a similar scale as the PASCAL VOC RGB dataset \cite{everingham2015pascal}. It also include all images data from NYU Depth v2 dataset \cite{silberman2012indoor}, and selected images data from Berkeley B3DO \cite{janoch2011category} and SUN3D \cite{xiao2013sun3d} dataset. To improve the quality of the depth map, the paper proposes a algorithm that estimates the 3D structure of the scene from multiple frames to conduct depth denoising and fill in the missing values. Each pixel in the RGB-D images is assigned a semantic label in one of the 37 classes or the `unknown' class. In the experiment evaluation, we use the default trainval-test split of the dataset that has 5285 training/validation instances and 5050 testing instances to evaluation our proposed RedNet architecture.

\textbf{Training}~ Images in SUN RGB-D dataset were captured by four different kinds of sensors with different resolutions and fields of view. In the training step, we resize all RGB images, Depth images, and the Groundtruth semantic maps into a \(480 \times 640\) height and width spatial resolution, additionally, the Groundtruth maps are further resized into four downsampled maps with resolution from \(240 \times 320\) to \(30 \times 40\) for pyramid supervision of the side output. Here, the RGB images are applied bilinear interpolation while the Depth images and Groundtruth maps are applied nearest-neighbor interpolation. During training, the inputs and Groundtruths data are augmented by applying random scale and crop and the input RGB images are further augmented by applying random hue, brightness, and saturation adjustment. In addition, we calculate the mean and standard deviation of the RGB and Depth images in the whole dataset to normalize each input value.

The two networks in the experiment, i.e., the RedNet (ResNet-34) and RedNet (ResNet-50), share the same training strategy and have the identical values of all hyperparameters. We use the PyTorch deep learning framework \cite{paszke2017automatic} for implementation and training of the architecture\footnote{Our source code will be avaliable at \url{https://github.com/JindongJiang/RedNet}}. The encoder of the network is pretrained on the ImageNet object classification dataset \cite{krizhevsky2012imagenet}, while the parameters on other layers are initialized by the Xavier initializer \cite{glorot2010understanding}. Since the imbalance of pixels of each class presented in the dataset, we reweight the training loss of each class in the cross-entropy function using the median frequency setting proposed in \cite{eigen2015predicting}. That is, we weight each pixel by a factor of \(\alpha_{c} = median\_prob/prob(c)\), where c is the groundtruth class of the pixel, \(prob(c)\) is the pixel probability of that class, \(median\_prob\) is the median of all the probabilities of these classes. The network is training with momentum SGD as the optimization algorithm. The initial learning rate of all layers are set to 0.002 and will decay by a factor of 0.8 in every 100 epochs. The momentum of the optimizer is set to 0.9, and a weight decay of 0.0004 is applied for regularization. The network is trained on a NVIDIA GeForce GTX 1080 GPU with a batch size of 5, and we stop the training when the loss no longer decrease.

\textbf{Evaluation}~ The network is evaluated on the default testing set of SUN RGB-D dataset. Three criterias for segmentation tasks are used to measure the performance of the network under 5050 testing instances, i.e., the pixel accuracy, the mean accuracy and the intersection-over-union (IoU) score.

\begin{table}[t]
    \small
    \caption{Comparison of SUN RGB-D testing results}
    \label{tab:testing_result}
    \centering
    \begin{tabular}{ l r r r }
        \toprule
        Model                    &~Pixel &~~Mean &~~mIoU     \\
        \midrule
        FCN-32s \cite{long2015fully} &68.4 &41.1 &29.0           \\
        SegNet \cite{badrinarayanan2015segnet} &71.2 &45.9 &30.7              \\
        Context-CRF \cite{lin2016exploring} &78.4 &53.4 &42.3             \\
        RefineNet-152 \cite{lin2017refinenet} &80.6 &58.5 &45.9           \\
        CFN (RefineNet-152) \cite{lin2017cascaded} &- &- &48.1    \\
        FuseNet-SF5 \cite{hazirbas2016fusenet} &76.3 &48.3 &37.3 \\
        DFCN-DCRF \cite{jiang2017incorporating} &76.6 &50.6 &39.3 \\
        \midrule
        RedNet(ResNet-34)                  &80.8 &58.3 &46.8  \\
        RedNet(ResNet-50)                  &\textbf{81.3} &\textbf{60.3} &\textbf{47.8} \\
        \bottomrule
    \end{tabular}
\end{table}

Table \ref{tab:testing_result} shows the comparison result of RedNet and other state-of-the-art methods on SUN RGB-D testing set. As we can see in the table, the proposed RedNet(ResNet-34) and RedNet(ResNet-50) architecture outperform most of the exist methods. Here, the FuseNet-SF5 \cite{hazirbas2016fusenet} and DFCN-DCRF \cite{jiang2017incorporating} networks use the same depth fusion technique in RedNet for depth incorporation. The RefineNet-152 \cite{lin2017refinenet} and CFN (RefineNet-152) \cite{lin2017cascaded} architecture use the same residual network in RedNet for feature extraction. Notice that, these two architectures are both using ResNet-152 structure for feature extraction, while RedNet performs a 47.8\% accuracy using the ResNet-50 as the encoder. It also worth notice that the RedNet(ResNet-34) network and the RedNet(ResNet-50) network share the same decoder structure, and the comparison result shows that the deeper structure of encoder in RedNet(ResNet-50) provides a better performance.

\begin{table}[t]
    \small
    \caption{SUN RGB-D testing results on pyramid supervision}
    \label{tab:pyramid_result}
    \centering
    \begin{tabular}{ l r r r }
        \toprule
        Model                    &~Pixel &~~Mean &~~mIoU     \\
        \midrule
      RedNet(ResNet-34) without pyramid   &80.3 &55.5 &45.0           \\
        RedNet(ResNet-34)                  &80.8 &58.3 &46.8  \\
      RedNet(ResNet-50) without pyramid  &80.5 &57.4 &46.0 \\
        RedNet(ResNet-50)                  &81.3 &60.3 &47.8 \\
        \bottomrule
    \end{tabular}
\end{table}

\begin{figure}[!t]
    \centering
    \includegraphics[width=0.7\textwidth,]{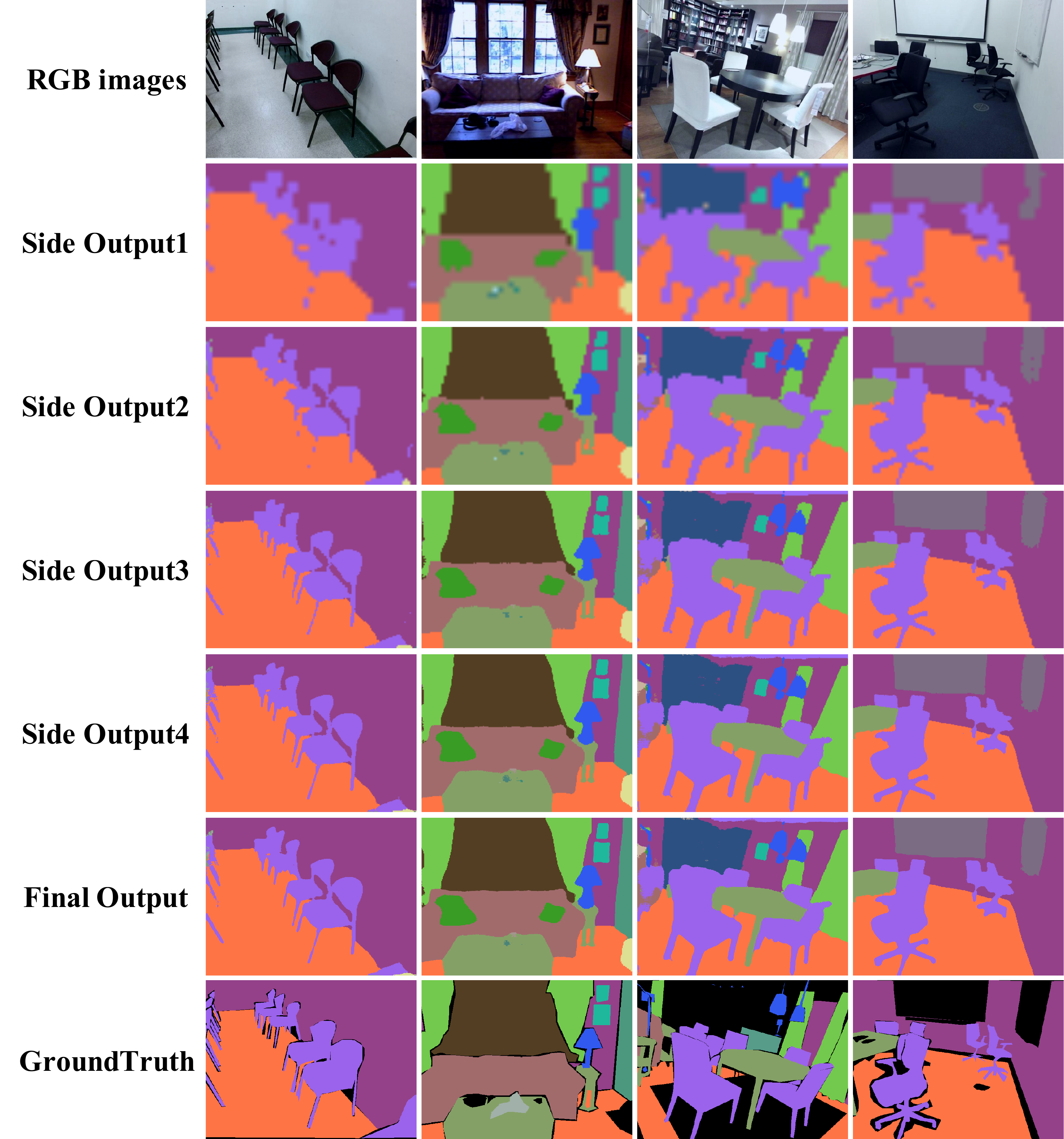}
    \caption{Prediction of side outputs and final output}
    \label{fig:outputs}
\end{figure}

In addition, to show that the pyramid supervision training scheme is able to effectively improve the performance of the network, a experiment is conducted to compare the performance of the proposed RedNet architectures trained with and without pyramid supervision. The result is shown in Table \ref{tab:pyramid_result}. It shows that the pyramid supervision improve the performance of the network on all three criterias. Notice that the ResNet-34 encoder RedNet with pyramid supervision training scheme outperform the ResNet-50 encoder RedNet without pyramid supervision, this fully demonstrate the effectiveness of pyramid supervision. The testing prediction of side outputs and final output can be obtained in Fig. \ref{fig:outputs}.

\section{Conclusion} \label{sec:conclusion}

In this work, we propose a RGB-D encoder-decoder residual network named RedNet for indoor RGB-D semantic segmentation. The RedNet combines the short skip-connection in residual unit and the long skip-connection between encoder and decoder for an accurate semantic inference. It also applies fusion structure in the encoder to incorporate the depth information. Moreover, we present the pyramid supervision training scheme that apply supervised learning over several layers on the decoder to improve the performance of the encoder-decoder network. The comparative experiment shows that the proposed RedNet architecture with pyramid supervision achieves state-of-the-art result on SUN RGB-D dataset.

\bibliographystyle{splncs04.bst}

\bibliography{reference}

\end{document}